\documentclass[a4paper,conference]{IEEEtran}
\usepackage{subfigure}
\usepackage{url}
\usepackage{times}
\usepackage{amssymb}
\usepackage{array}
\usepackage{multirow}
\usepackage[ruled,lined]{algorithm2e}
\usepackage{amsmath}
\usepackage{booktabs}
\usepackage{cite}
\usepackage{caption}
\ifCLASSINFOpdf
  \usepackage[pdftex]{graphicx}

  \graphicspath{{../pdf/}{../jpeg/}}

  \DeclareGraphicsExtensions{.pdf,.jpeg,.png}
\else

  \usepackage[dvips]{graphicx}
  \graphicspath{{../eps/}}
  \DeclareGraphicsExtensions{.eps}
\fi

\begin{document}

\title{\large \bf A DISCRIMINATIVELY LEARNED CNN EMBEDDING FOR REMOTE SENSING IMAGE SCENE CLASSIFICATION \\}

\author{
Wen Wang$^1$, Lijun Du$^2$, Yinxing Gao$^1$, Yanzhou Su$^1$, Feng Wang$^1$, Jian Cheng$^1$\\
${}^1$University of Electronic Science and Technology of China,\\
School of Information and Communication Engineering\\
${}^2$Leshan Normal University, School of Computer Science\\
uestc\_wangwen@163.com, chengjian@uestc.edu.cn }

\maketitle
\begin {abstract}
In this work, a discriminatively learned CNN embedding is proposed for remote sensing image scene classification. Our proposed siamese network simultaneously computes the classification loss function and the metric learning loss function of the two input images. Specifically, for the classification loss, we use the standard cross-entropy loss function to predict the classes of the images. For the metric learning loss, our siamese network learns to map the intra-class and inter-class input pairs to a feature space where intra-class inputs are close and inter-class inputs are separated by a margin. Concretely, for remote sensing image scene classification, we would like to map images from the same scene to feature vectors that are close, and map images from different scenes to feature vectors that are widely separated. Experiments are conducted on three different remote sensing image datasets to evaluate the effectiveness of our proposed approach. The results demonstrate that the proposed method achieves an excellent classification performance.\footnotetext{W. Wang, Y. Gao, Y. Su, F. Wang and J. Cheng are with the School of Information and Communication Engineering, University of Electronic Science and Technology of China, Chengdu, Sichuan, China, 611731. L. Du is with the school of computer science, University of Leshan Normal. This research has been supported by the National Natural Science Fundation of China under Grant 61671125. The Fundamental Research Funds for the Central Universities NO.2672018ZYGX2018J008. (Corresponding author: chengjian@uestc.edu.cn)}
\end{abstract}

\IEEEpeerreviewmaketitle

\section{Introduction}

With the development of remote sensing technology, large amounts of high-resolution images are becoming increasingly available and remote sensing image classification has attracted much attention in recent years\cite{Zhang2014Saliency},\cite{Hu2015Transferring},\cite{Zou2015Deep}. Convolutional Neural Network (CNN) has shown great potential for remote sensing image scene classification. However, scene classification is still a challenging problem because of the large variability of scale, orientation, illumination, viewpoint and layout of scene in the images.

\begin{figure}[h]
\centering
\includegraphics[width=9cm]{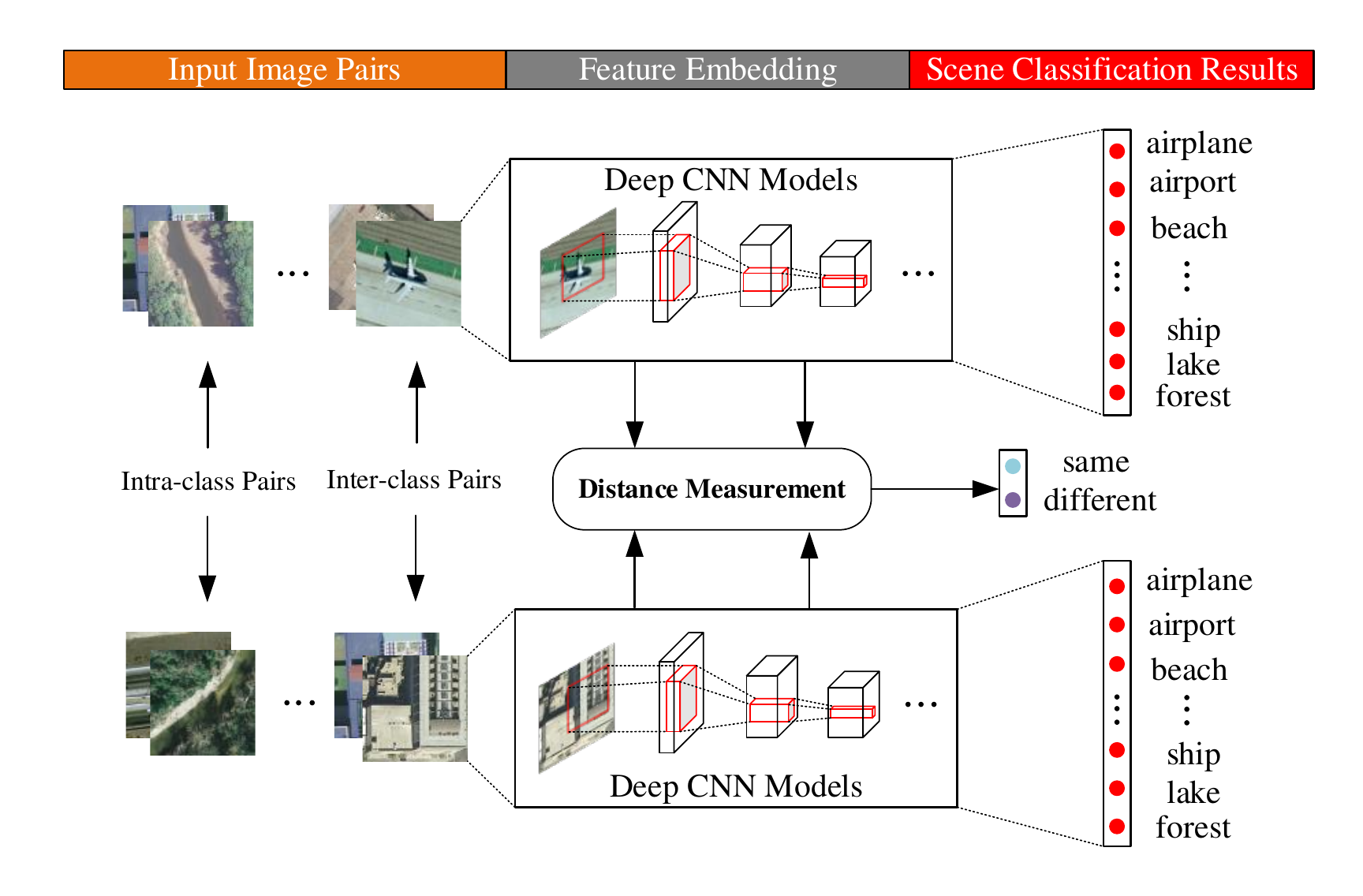}
\caption{The structure of our siamese network: a pair of scene images are fed into two deep CNN models for feature embedding. And then, the output two feature embeddings are used to predict the classes of the two input images, respectively, and also measure the distance jointly. Finally, the network optimizes the three objectives.  }
\label{fig_1}
\end{figure}

During the past few years, there has been increased interest in developing a variety of methods to deal with remote sensing image scene classification. Traditional methods design the hand-craft features, such as color, texture and shape, including SIFT\cite{Lowe2004}, LBP\cite{1017623}, HOGs\cite{dalal:inria-00548512} and Gist \cite{Oliva2001Modeling}, which are the primary characteristics of a scene image. Then, the descriptors based on visual dictionaries (e.g., Bag of Visual Words (BoVW) model)\cite{Zhu2017Bag}, attracted the attention and were widely used for feature encoding, in which the input is a set of handcrafted features and the output is a set of learned features. In recent years, with the development of deep learning techniques, especially Convolutional Neural Network (CNN), deep feature learning-based methods have shown high feature representativeness and generalization capability for remote sensing image scene classification\cite{Nogueira2017Towards}, \cite{Cheng2017Remote}, \cite{Penatti2015Do}. In deep learning-based methods, a robust and discriminative descriptor is usually employed to represent each input image as a feature vector, where different scene classes are expected to be separated as much as possible in the feature space. However, this model has a relatively weak constraint on features extracted from the same scene, since dissimilar features of the same scene classes would be mapped to different scene classes, which lead to low scene classification accuracy.

In this paper, we use metric learning loss function to learn from the labeled training images to effectively measure the distance of scene samples, under which the distance of inter-class input pairs are enlarged and that of intra-class input pairs are reduced as much as possible to improve the discriminative ability of the learned embedding for remote sensing image classification. The proposed network is a siamese network architecture \cite{Hadsell2006Dimensionality} that predicts scene classes and measures the distance at the same time. Compared to the previous models, we take full advantages of the labeled training data in terms of pair-wise distance similarity and image classes. To summarize, our contributions are two-fold:

(1) We propose to use metric learning loss function to learn a discriminative CNN embedding by the siamese loss, which learns the class for each scene and penalizes the distance between the deep features and their corresponding labels. 

(2) Our proposed method significantly outperforms the state-of-art methods on three remote sensing datasets: Brazilian Coffee Scenes \cite{Penatti2015Do}, UCMerced LandUse \cite{Yang2010Bag} and NWPU-RESCISC45 \cite{Cheng2017Remote}.

The rest of this paper is organized as follows. Section II introduces the proposed method in detail. Section III presents the experimental results on three publicly available scene datasets. At last, we conclude our works in Section IV.

\section{Our approach}
The goal of our proposed method is to extract the features from training images and compute the distance between images with a discriminative metric for accurate image classification. Fig.\ref{fig_1} shows an overview of the siamese network for remote sensing image scene classification. 

\subsection{Overall Network}
Our network is basically a convolutional siamese network, which consists of two sub-networks with shared weights. Given an input pair of scene images, the proposed network simultaneously predicts the classes and the distance measurement of the two input images.

As shown in Fig.\ref{fig_1}, the images are processed by a Convolutional Neural Network (CNN). The CNN involves many individual processing steps, so we refer to the complete CNN as a function, $f=C(x,\theta_{c})$, that takes an image $x$ as input and produces a vector $f$ as output, where $f$ is the vectorised representation of the CNN's final layer activation maps and $\theta_{c}$ denotes ConvNets parameters to be learned. In this work, we use GoogleNet\cite{Szegedy2014Going} as the base CNN architecture, which was pre-trained on ImageNet \cite{Olga2015ImageNet}.  

\subsection{Siamese Loss}
Our siamese network trains the feature extraction network to optimize both the classification loss to predict the scene classes and the distance learning loss to estimate similarity. 

The first is the classification loss function, which classifies remote sensing images into one of $n$ different classes. The classification loss function is achieved by following the deep CNN with an $n$-way softmax layer, which outputs a probability distribution over the $n$ classes. The network is trained to minimize the softmax function, or cross-entropy loss, which is denoted as,
\begin{equation}
    z_{i} = W_{i}^{T}f+b_{i},
\end{equation}
\begin{equation}
   P_{c}=P(q=c|f)=\frac{exp(z_{c})}{\sum _{k=1}^{n}exp(z_{k})},
\end{equation}

\begin{equation}\label{eq_1}
I(f,\theta_{i})=-logP_{c},
\end{equation}
where $W$ is the softmax weight matrix, $b$ is the bias matrix, and $c$ is the target class, $P_{c}$ is the predicted probablity, and $f$ is the image feature vector, $\theta_{i}$ denotes the softmax layer parameters.

The second is the distance learning loss function, which encourages features extracted from scene images of the same classes to be similar and enlarges the margin between the features from different classes in the feature space. Given a pair of images ($s_{i},s_{j}$), where each image has been processed using the deep CNN feature extraction network to give image feature vectors, $f_{i}=C(s_{i})$ and $f_{j}=C(s_{j})$, where $C(\cdot)$ is the feature extraction function defined by the CNN. The high-level feature from the fine-tuned CNN has shown a discriminative ability and it is more compact than the activations in the intermediate layers\cite{Chen2014Deep}. So we directly compares the high-level features $f_{i}$, $f_{j}$ for the distance estimation. We can write the siamese network training objective as a function of the feature vectors $f_{i}$ and $f_{j}$ as follows,
\begin{equation}\label{eq_2}
V(f_{i},f_{j},\theta_{v})=\left\{\begin{matrix}
\frac{1}{2}||f_{i}-f_{j}||_{2}^{2} & i=j,\\ 
 \frac{1}{2}[max(m-||f_{i}-f_{j}||_{2}｝,0)]^{2}& i\neq j.
\end{matrix}\right.
\end{equation}
where $||f_{i}-f_{j}||_{2}^{2}$ is the Euclidean distance between the feature vectors. When two images are from the same scene classes ($i=j$), the objective $V$ encourages the features $f_{i}$ and $f_{j}$ to be close by minimizing the $L_{2}$ distance between the two vectors. While for images from different scene classes ($i\neq j$), it encourages the distance larger than a margin $m$. $\theta_{v}$ is the parameter to be learned in the above training objective function.

We can now define the overall training objective $L$ for a single pair of images($s_{i},s_{j}$), which jointly optimizes the classification loss function and the distance learning loss function to train the CNN for discriminative feature learning. The formulation is given as follows,
\begin{equation}\label{eq_3}
    L(s_{i},s_{j})=I(C(s_{i}))+I(C(s_{j}))+\lambda V(C(s_{i}),C(s_{j})),
\end{equation}
where $\lambda$ is used for balancing the two loss functions. 

\subsection{Optimization}

Our goal is to learn the parameters $\theta_{c}$ in the feature extraction function $C$, while $\theta_{i}$ and $\theta_{v}$ are only parameters introduced to propagate the classification loss and the distance learning loss during the training stage. In the testing stage, only $\theta_{c}$ is used for feature extraction. The parameters are updated by stochastic gradient descent. 

Then the gradient of $L$ with respect to $f_{i}$ and $f_{j}$ are given by,
\begin{equation}
    \frac{\partial L}{\partial f_{i}} = \frac{\partial I(f_{i},\theta_{i})}{\partial f_{i}} + \lambda \frac{\partial V(f_{i},f_{j},\theta_{v})}{\partial f_{i}},
\end{equation}
\begin{equation}
    \frac{\partial L}{\partial f_{j}} = \frac{\partial I(f_{j},\theta_{i})}{\partial f_{j}} + \lambda \frac{\partial V(f_{i},f_{j},\theta_{v})}{\partial f_{j}}. 
\end{equation}

The gradients of $V$ with respect to $f_{i}$ is defined by the following relation, and the gradients of $V$ with respect to $f_{j}$ is symmetric with that of $f_{i}$. The derivatives are used to update the parameters of deep nueral networks. 

\begin{equation}\label{eq_6}
    \left\{\begin{matrix}
 \frac{\partial V}{\partial f_{i}}=\left \| f_{i}-f_{j} \right \|_{2}\frac{\partial \left \| f_{i}-f_{j} \right \|_{2}}{\partial f_{i}}=f_{i}-f_{j} & i=j,\\ 
 \frac{\partial V}{\partial f_{i}}=-max[(m-\left \| f_{i}-f_{j} \right \|_{2}),0]\frac{f_{i}-f_{j}}{\left \| f_{i}-f_{j} \right \|_{2}}&i\neq j. 
\end{matrix}\right.
\end{equation}

The gradients of $I$ with respect to $z_{i}$ is illustrated as follows,

\begin{equation}\label{eq_10}
\frac{\partial I}{\partial z_{i}}=\left\{\begin{matrix}
P_{i}-1 & i=c, \\ 
 P_{i}&i\neq c.
\end{matrix}\right.
\end{equation}

At last, the network updates the parameters of $\theta_{i}$, $\theta_{v}$ and $\theta_{c}$ as,  $\theta_{i}=\theta_{i}-\eta \cdot \frac{\partial L}{\partial \theta_{i}}$, $ \theta_{v}=\theta_{v}-\eta \cdot \frac{\partial L}{\partial \theta_{v}} $
and 
    $ \theta_{c}=\theta_{c}-\eta \cdot \frac{\partial L}{\partial \theta_{c}}$, where $\eta$ is the learning rate.

\section{Experiments}
In order to better evaluate the effectiveness of our proposed model, we have chosen three remote sensing datasets with different visual properties. The details about the three datasets and the experiments implementation are presented in the following subsections. Finally, we present and discuss the experimental results.

\begin{figure}[h]
\centering
\includegraphics[width=9cm]{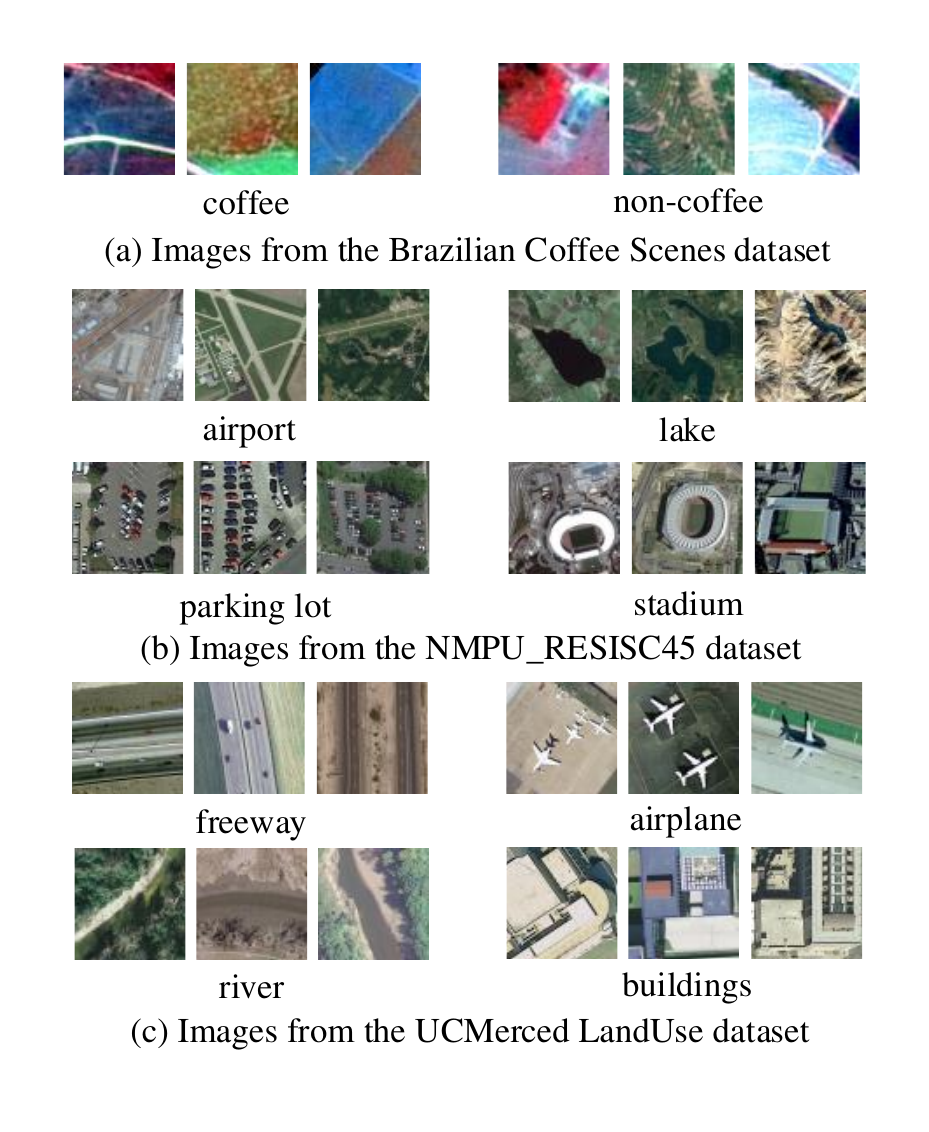}
\caption{Some example images from the three datasets.}
\label{fig_2}
\end{figure}

\subsection{Dataset}
\noindent \textbf{UCMerced LandUse.} The UCMerced LandUse dataset \cite{Yang2010Bag} is one of the first publicly avaliable high-resolution remote sensing imagery datasets. This dataset contains $2100$ aerial scene images with $256\times 256$ pixels equally divided into $21$ land-use classes. Fig.\ref{fig_2} (c) shows some examples of ground truth images from three classes in this dataset.

\noindent \textbf{Brazilian Coffee Scenes.} This dataset \cite{Penatti2015Do} includes multi-spectral scenes taken by the SPOT sensor. It contains $2876$ images with $64\times 64$ pixels equally divided into $2$ classes (coffee and non-coffee). Fig.\ref{fig_2} (a) shows some examples of this dataset. It has many intra-class variances caused by different crop management techniques.  

\noindent \textbf{NWPU-RESISC45.} This dataset \cite{Cheng2017Remote} contains $31500$ images, covering $45$ scene classes. Each class consists of $700$ images with the size of $256\times256$ pixels. Fig.\ref{fig_2} (b) shows some samples of two classes from this dataset. This dataset is one of the largest scale on the number of the scene classes and the total number of images.

\subsection{Implementation details}

\begin{table}[]
\centering
\caption{Classification accuracy achieved on three different datasets.}
\label{table_1}
\begin{tabular}{p{2cm}<{\centering}|p{3cm}<{\centering}|p{1.5cm}<{\centering}}
\hline
\textbf{Dataset} & \textbf{Method} & \textbf{Accuracy} \\ \hline
\multirow{2}{*}{\begin{tabular}[c]{@{}c@{}}UCMerced \\ LandUse\end{tabular}} & GoogleNet-Basel.\cite{Nogueira2017Towards} & 95.47\% \\ \cline{2-3} 
 & Ours(GoogleNet) & {\bf96.26\%} \\ \hline
\multirow{2}{*}{\begin{tabular}[c]{@{}c@{}}Brazilian\\ Coffee\end{tabular}} & GoogleNet-Basel.\cite{Nogueira2017Towards} & 92.11\% \\ \cline{2-3} 
 & Ours(GoogleNet) & {\bf93.65\%} \\ \hline
\multirow{2}{*}{\begin{tabular}[c]{@{}c@{}}NWPU-\\ RESISC45\end{tabular}} & GoogleNet-Basel.\cite{Cheng2017Remote} & 86.02\% \\ \cline{2-3} 
 & Ours(GoogleNet) & {\bf89.16\%} \\ \hline
\end{tabular}
\end{table}

For UCMerced LandUse dataset, all the images are randomly cropped to $227\times227$ and mirrored horizontally during training. For Brazillian Coffee Scenes and NWPU-RESISC45 dataset, we randomly crop images to $64\times 64$ and $227\times 227$ respectively. The mean image computed from all the training images is substracted from all the images. Besides, we shuffle the dataset and use a random order of the images. The margin $m$ in our experiment is set to $1$. All experiments were performed on a 64 bits Intel i7-4790 machine with 32GB of RAM memory. A GeForce GTX 1080Ti with 11GB of memory and Ubuntu version 14.04.1 LST were used in our experiments.

\subsection{Classification Evaluation}
\noindent {\bf Comparison with the CNN baseline.} We trained the baseline networks without the distance measurement. The baseline nteworks were pretrained on ImageNet and fine-tuned to predict the scene classes. As shown in TABLE \ref{table_1}, the classification accuracy can achieve $ 96.26\%$, $ 93.65\%$ and $ 89.16\%$ on UCMerced LandUse, Brazilian coffee and NWPU-RESISC45 dataset seperately which are better than $95.47\%$, $92.11\%$ and $86.02\%$ obtained by the baseline network on GoogleNet. The results show that: (1) distance measurement used in remote sensing image has positive effects on improving the performance of classification accuracy. (2) our proposed method can work with different datasets and improve their results. (3) the proposed model helps the network to learn more discriminative features.

\begin{figure}[h]
 \centering
\includegraphics[width=8.5cm]{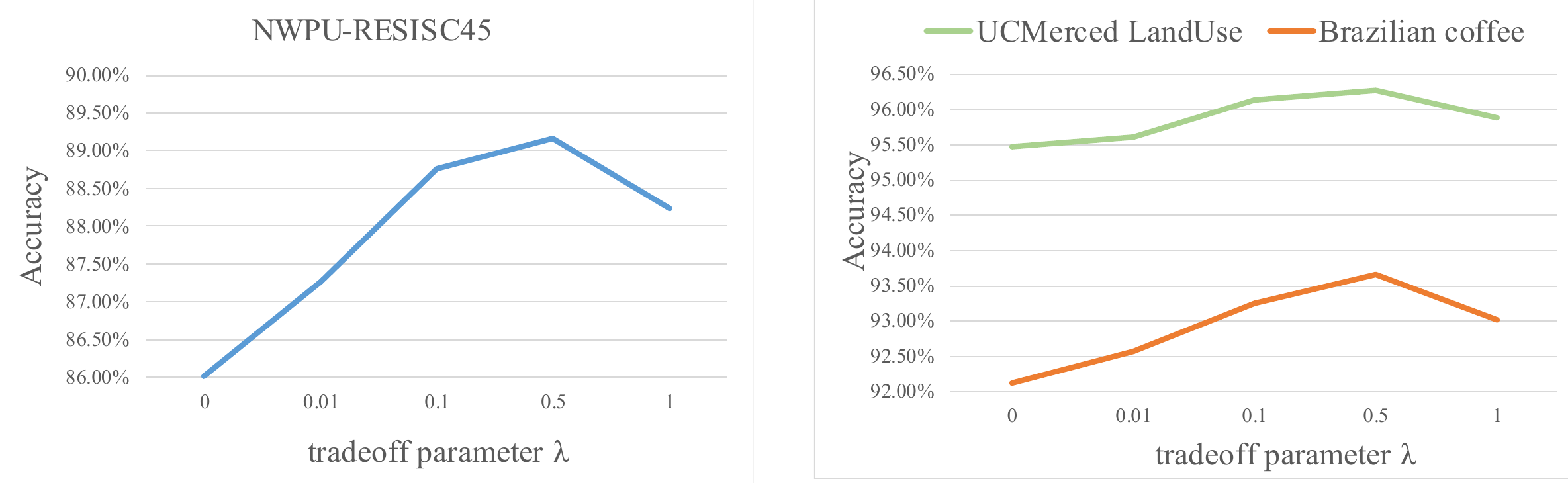}
\caption{The effects of tradoff parameter $\lambda$ on accuracy.}
\label{fig_6}
\end{figure}

\noindent {\bf Comparison with different values of $\lambda$.}
$\lambda$ is a tradeoff parameter to balance the contribution of the classification loss and the distance learning loss. The value of $\lambda$ affects the classification accuracy. As shown in Fig.\ref{fig_6}, the results show that: (1) distance measurement has positive effects on improving the classification accuracy. (2) a larger $\lambda$ means higher accuracy, while excessively larger $\lambda$ decreases the performance. For the reason that, if $\lambda$ is too large, too much attention is paid to the distance measurement, and the classification prediction is ignored.

\begin{figure}[h]
\centering
\includegraphics[width=8cm]{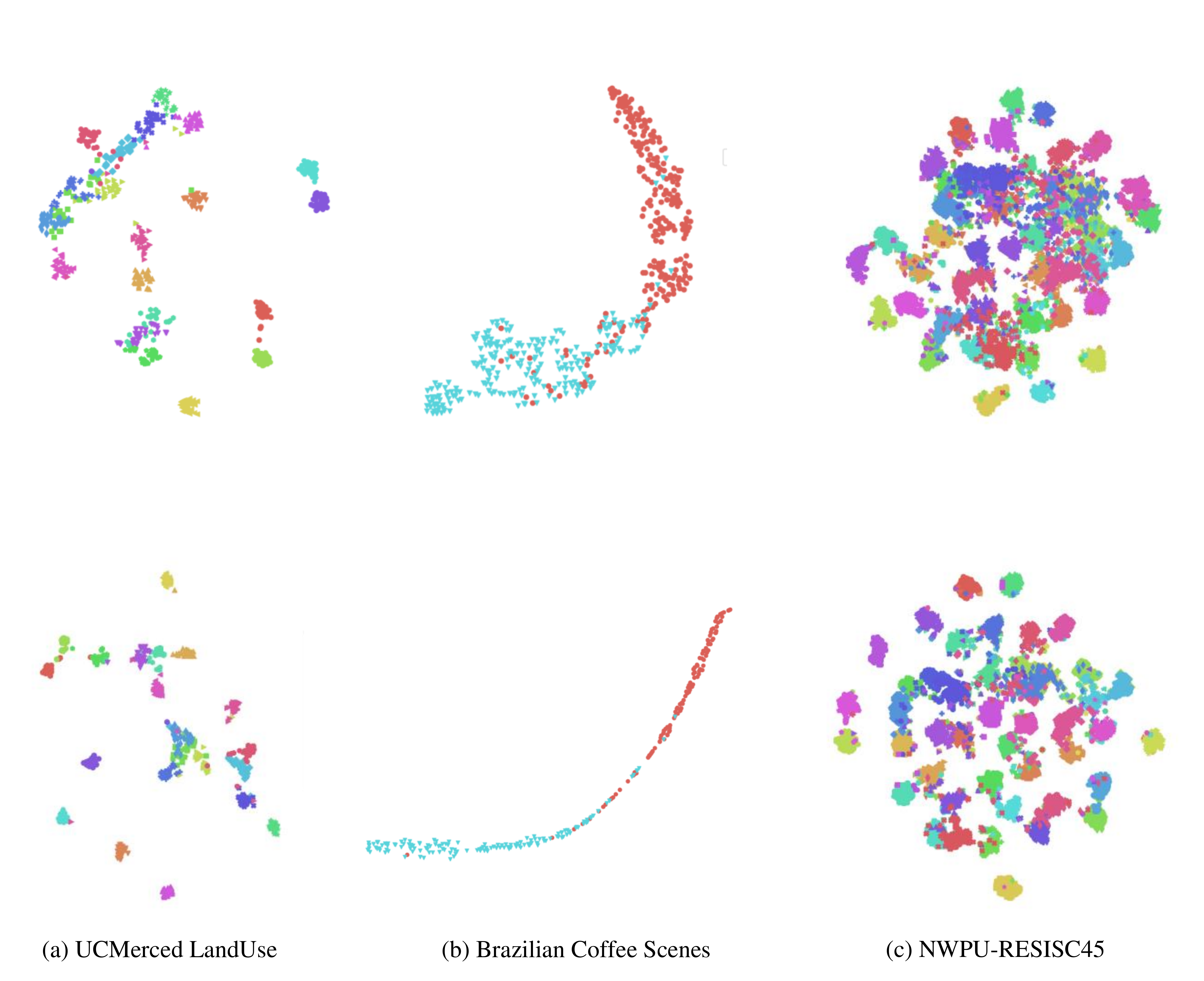}
\caption{Feature embedding visualizations of the baseline (first line) and our proposed embedding (second line) on the test splits of the three scene datasets using t-SNE\cite{Maaten2008VisualizingDU}.}
\label{fig_5}
\end{figure}

\noindent {\bf Feature embedding.} We qualitatively evaluate our learned features to verify if it is a good generic feature. We extract the fully-connected layer features from test split of the three datasets. These features are then projected to 2-dimensional space using t-SNE\cite{Maaten2008VisualizingDU}. Fig.\ref{fig_5} shows the feature visualization on the learned embedding of our model and the baseline CNN. Under the single classification signal of cross-entropy loss, the features are less seperated than our proposed method with both classification loss and distance metric learning loss. If we only use the softmax loss, the resulting deeply learned features would contain large intra-class variations. While joint the softmax loss and the distance metric learning loss achieve good intra-class compactness and inter-class separability, which is very beneficial to the discriminative feature learning.

\section{Conclusion}
In this paper, we have proposed to combine class prediction and distance measurement to improve the performance for remote sensing image scene classification. Our proposed method takes full advantage of the similarity information in training samples and learns a discriminative embedding. It outperforms the CNN baseline on three different remote sensing datasets and shows the effectiveness of the proposed method on improving the remote sensing image classification performance. 

\bibliographystyle{IEEEtran}

\bibliography{reference}

\begin{thebibliography}{10}

\bibitem{Chen2016Land}
Chen Chen, Baochang Zhang, Hongjun Su, Wei Li, and Lu~Wang.
\newblock Land-use scene classification using multi-scale completed local
  binary patterns.
\newblock {\em Signal Image \& Video Processing}, 10(4):745--752, 2016.

\bibitem{Chen2014Deep}
Yuheng Chen, Yuheng Chen, Xiaogang Wang, and Xiaoou Tang.
\newblock Deep learning face representation by joint
  identification-verification.
\newblock In {\em International Conference on Neural Information Processing
  Systems}, pages 1988--1996, 2014.

\bibitem{Cheng2017Remote}
Gong Cheng, Junwei Han, and Xiaoqiang Lu.
\newblock Remote sensing image scene classification: Benchmark and state of the
  art.
\newblock {\em Proceedings of the IEEE}, 105(10):1865--1883, 2017.

\bibitem{dalal:inria-00548512}
Navneet Dalal and Bill Triggs.
\newblock {Histograms of Oriented Gradients for Human Detection}.
\newblock In Cordelia Schmid, Stefano Soatto, and Carlo Tomasi, editors, {\em
  {International Conference on Computer Vision \& Pattern Recognition (CVPR
  '05)}}, volume~1, pages 886--893, San Diego, United States, June 2005. {IEEE
  Computer Society}.

\bibitem{Hadsell2006Dimensionality}
R.~Hadsell, S.~Chopra, and Y.~Lecun.
\newblock Dimensionality reduction by learning an invariant mapping.
\newblock In {\em IEEE Computer Society Conference on Computer Vision and
  Pattern Recognition}, pages 1735--1742, 2006.

\bibitem{Hu2015Transferring}
Fan Hu, Gui~Song Xia, Jingwen Hu, and Liangpei Zhang.
\newblock Transferring deep convolutional neural networks for the scene
  classification of high-resolution remote sensing imagery.
\newblock {\em Remote Sensing}, 7(11):14680--14707, 2015.

\bibitem{Lowe2004}
David~G. Lowe.
\newblock Distinctive image features from scale-invariant keypoints.
\newblock {\em International Journal of Computer Vision}, 60(2):91--110, Nov
  2004.

\bibitem{Nogueira2017Towards}
Keiller Nogueira, Otávio A.~B. Penatti, and Jefersson A.~Dos Santos.
\newblock Towards better exploiting convolutional neural networks for remote
  sensing scene classification.
\newblock {\em Pattern Recognition}, 61:539--556, 2017.

\bibitem{Oliva2001Modeling}
Aude Oliva and Antonio Torralba.
\newblock Modeling the shape of the scene: A holistic representation of the
  spatial envelope.
\newblock {\em International Journal of Computer Vision}, 42(3):145--175, 2001.

\bibitem{Penatti2015Do}
Otavio A.~B. Penatti, Keiller Nogueira, and Jefersson A.~Dos Santos.
\newblock Do deep features generalize from everyday objects to remote sensing
  and aerial scenes domains?
\newblock In {\em IEEE Conference on Computer Vision and Pattern Recognition
  Workshops}, pages 44--51, 2015.

\bibitem{Olga2015ImageNet}
Olga Russakovsky, Jia Deng, Hao Su, Jonathan Krause, Sanjeev Satheesh, Sean Ma,
  Zhiheng Huang, Andrej Karpathy, Aditya Khosla, and Michael Bernstein.
\newblock Imagenet large scale visual recognition challenge.
\newblock {\em International Journal of Computer Vision}, 115(3):211--252,
  2015.

\bibitem{Szegedy2014Going}
Christian Szegedy, Wei Liu, Yangqing Jia, Pierre Sermanet, Scott Reed, Dragomir
  Anguelov, Dumitru Erhan, Vincent Vanhoucke, and Andrew Rabinovich.
\newblock Going deeper with convolutions.
\newblock pages 1--9, 2014.

\bibitem{Maaten2008VisualizingDU}
Laurens van~der Maaten and Geoffrey~E. Hinton.
\newblock Visualizing data using t-sne.
\newblock 2008.

\bibitem{Yang2010Bag}
Yi~Yang and Shawn Newsam.
\newblock Bag-of-visual-words and spatial extensions for land-use
  classification.
\newblock In {\em Sigspatial International Conference on Advances in Geographic
  Information Systems}, pages 270--279, 2010.

\bibitem{Zhang2014Saliency}
Fan Zhang, Bo~Du, and Liangpei Zhang.
\newblock Saliency-guided unsupervised feature learning for scene
  classification.
\newblock {\em IEEE Transactions on Geoscience \& Remote Sensing},
  53(4):2175--2184, 2014.

\bibitem{Zhu2017Bag}
Qiqi Zhu, Yanfei Zhong, Bei Zhao, Gui~Song Xia, and Liangpei Zhang.
\newblock Bag-of-visual-words scene classifier with local and global features
  for high spatial resolution remote sensing imagery.
\newblock {\em IEEE Geoscience \& Remote Sensing Letters}, 13(6):747--751, 2017.

\bibitem{Zou2015Deep}
Qin Zou, Lihao Ni, Tong Zhang, and Qian Wang.
\newblock Deep learning based feature selection for remote sensing scene
  classification.
\newblock {\em IEEE Geoscience \& Remote Sensing Letters}, 12(11):2321--2325,
  2015.

\end{thebibliography}

\end{document}